\documentclass[conference]{aiaa-pretty}
\usepackage[utf8]{inputenc}
\usepackage[english]{babel}
\usepackage{graphicx}
\usepackage{subcaption}
\usepackage{caption}
\usepackage{listings}
\usepackage{amsmath}
\usepackage{multirow}
\usepackage[linesnumbered,ruled,vlined]{algorithm2e}

\usepackage{xcolor}

\author[Jerry]{ %
Jun Xiang\thanks{Ph.D. Student, Department of Aerospace Engineering, AIAA Student Member, \texttt{jxiang9143@sdsu.edu }},
Junfei Xie \thanks{Associate Professor, Department of Electrical and Computer Engineering, AIAA Member, \texttt{jxie4@sdsu.edu}},
Jun Chen\thanks{Assistant Professor, Department of Aerospace Engineering, AIAA Senior Member, \texttt{Jun.Chen@sdsu.edu}}\\
\textit{San Diego State University, San Diego, CA 92182}}

\title{Learning-accelerated A* Search for Risk-aware Path Planning }
 \abstract{
Safety is a critical concern for urban flights of autonomous Unmanned Aerial Vehicles. In populated environments, risk should be taken into account to produce an effective and safe path, known as risk-aware path planning. Risk-aware path planning can be modeled as a Constrained Shortest Path (CSP) problem, aiming to identify the shortest possible route that adheres to specified safety thresholds. CSP is NP-hard and poses significant computational challenges. Although many traditional methods can solve it accurately, all of them are very slow. Our method introduces an additional safety dimension to the traditional A* (called ASD A*), enabling A* to handle CSP. Furthermore, we develop a custom learning-based heuristic using transformer-based neural networks, which significantly reduces the computational load and improves the performance of the ASD A* algorithm. The proposed method is well-validated with both random and realistic simulation scenarios.
}

\begin{document}

\maketitle

\section{Introduction}

Injuries and property damage resulting from drones in populated areas are not unusual~\cite{plioutsias2018hazard}. Safety is the most critical factor to operate Unmanned Aerial Vehicles (UAVs) in the airspace while efficiency is also essential. Therefore, the risk associated with flying over populated areas must be addressed when planning the flight path and the planned path should cost minimal resources. Drone traffic is expected to increase due to the delivery of items and food as well as the transportation of people. In recent years, research on the planning of flight paths for drones has been conducted in several ways~\cite{guo2021safety}. The goal of this kind of research is to find the flight path with the lowest cost while avoiding high-risk areas, such as roads and highways~\cite{schopferer2020minimum}.

Finding the shortest path with safety constrained is a Constrained Shortest Path (CSP) problem. The CSP problem is NP-hard even for acyclic networks~\cite{wang1996quality}.
Currently, the CSP problem can be solved by many methods. Dynamic programming via the labeling of nodes is one of the earliest approaches to this problem~\cite{desrochers1988generalized}. A lot of dynamic programming-based method is proposed \cite{marinakis2017hybrid, resende2014grasp, tilk2017asymmetry, wu2020probabilistically}. Lagrangian Relaxation \cite{handler1980dual} of the weight constraint is another important approach. Kth-shortest path~\cite{dumitrescu2003improved}~\cite{eppstein1998finding} method is used to close the duality gap obtained from solving a Lagrangian relaxation. However, all of those methods are very slow even with the accelerating algorithm. For a complicated problem, the traditional solver may take days.

For fast computing, Heuristic approaches and approximate algorithms have been proposed~\cite{xiang2023hybrid}. Heuristic methods typically offer no assurance regarding the quality of the solutions they generate, even though they tend to operate quickly in practical scenarios. In contrast, $\epsilon$-approximation algorithms yield solutions that have a cost at most (1 + 
$\epsilon$) times the optimal cost. However, these algorithms often work slowly in practical situations due to their commitment to ensuring the quality of the solutions they provide~\cite{xiao2005constrained}.
Graph searching is another popular method to solve the path planning of drones. In fully visible airspace, searching for the shortest path from a start node to a goal node is easy. Nevertheless, path planning of drones considers not only the distance of the path but also many other features such as efficiency, schedule, and most importantly safety. One similar example is the vehicle routing problem with time window~\cite{irnich2005shortest,chen2018probabilistic}.  In real airspace, we need a large amount of nodes and edges to model the airspace. Therefore, a faster solver is urgently needed.

Imitation learning has been developed over more than 30 years. In 1988, Pomerleau trained a Deep Neural Network with expert actions to drive an autonomous vehicle with Behavior Cloning~\cite{pomerleau1988alvinn}. Recent research shows that Inverse Reinforcement Learning (IRL) outperforms Behavior Cloning~\cite{ghasemipour2020divergence}. IRL can output a policy to attain performance close to that of the expert without the reward information~\cite{abbeel2004apprenticeship}. Generative Adversarial Imitation Learning~\cite{ho2016generative} can also imitate experts who perform complex behaviors in large, high-dimensional environments. This method is also extended to apply to the multi-agent problem~\cite{song2018multi}.  Learning based method has also been approved to solve the combinatorial optimization problem~\cite{bengio2021machine}. Some datasets~\cite{tang2022new} is created to become the expert for Imitation learning.

Deep Learning has been one of the most popular research topics due to the increase in computing capacity and demand for artificial intelligence. Deep Learning can mimic experts' behavior and finish the given task~\cite{hussein2017imitation}.
Unsupervised learning algorithms train the agent from unlabeled data, making it ideal for exploratory analysis and discovering hidden patterns in data. Unsupervised learning algorithms can be applied to a wide range of applications, including object detection~\cite{chen2021multimodal}, image and speech recognition, recommendation systems, and more. 
One recent example of success of the unsupervised learning is the Generative Pre-trained Transformer (GPT).
GPT proposed by OpenAI is based on the Transformer architecture, which was introduced in the paper "Attention is All You Need" by Vaswani et al.~\cite{vaswani2017attention}. The Transformer architecture is designed to handle sequences of data, such as text or speech, and is well suited for natural language processing (NLP) tasks. GPT is notable for its remarkable performance in many NLP tasks, as well as its ability to generate high-quality text, images, and even music. It has become one of the most widely used deep learning models in the field of NLP and has had a significant impact on the development of AI.

This paper proposes to accelerate the A* search with transformer-based neural networks for NP-hard risk-aware path planning. The major contributions of this paper are summarized as follows:
\begin{itemize}
    \item This paper introduces an additional safety dimension to the A* algorithm, called ASD A*, which can help find the constrained shortest path for risk-aware path planning. 
    \item 
    This paper develops a custom learning-based heuristic using transformer-based neural networks, which significantly reduces the computational load and improves the performance of the ASD A* algorithm.
\end{itemize}

The paper's structure is as follows. Section \ref{sec:ps} introduces the problem setup. Section~\ref{sec:literature} introduced the previous method to estimate the heuristic for A*; Section~\ref{sec: method} introduces the additional safe dimension A*, how the expert heuristic is created, and the neural network architecture; Section~\ref{sec: results} shows the effectiveness and efficiency of the accelerated A*; Section~\ref{sec: conc} concludes this paper.

\section{Problem Statement}\label{sec:ps}
\begin{figure}[!htb]
 \centering
 \includegraphics[width=0.5\linewidth]{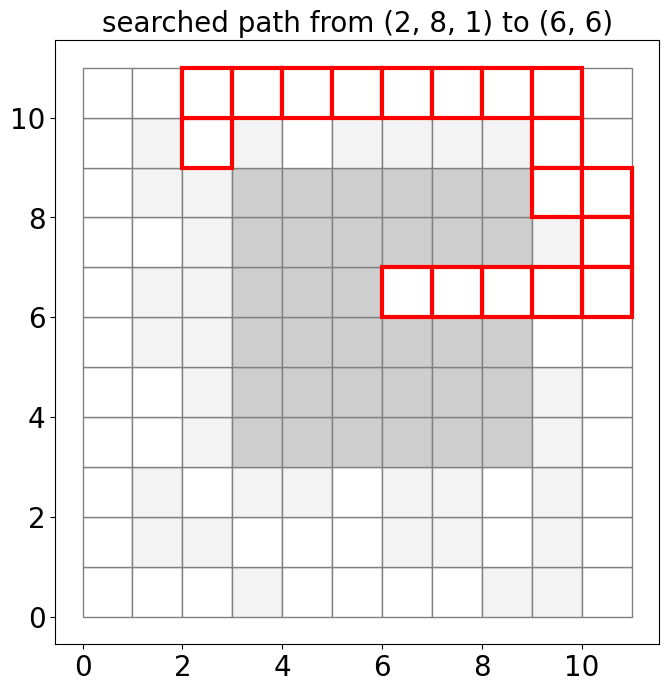}
 \captionof{figure}{an 11*11 risk map example, where grayscale represents risk value.}
  \label{fig:risk}
\end{figure}

In this paper, the ultimate target is to find the risk-constrained shortest path on a risk map of airspace.
To make A* available, the airspace first needs to be discretized into a grid world as shown in Fig~\ref{fig:risk}.
After discretization, the airspace is represented by a $m \times m$ grid graph $G$ and a set of edges $E$ which is defined as follows:
\begin{equation}
    G = \{ g_{xy} \}_{m \times m}  \quad where: x \in \{1, 2, ...,m\}, y \in \{1, 2, ...,m\},
\end{equation}
\begin{equation}
    E = \{ e_{u v}=(u, v) \}  \quad where: u \in \mathbf{Adj}(v), v \in \mathbf{Adj}(u), \forall u, v \in G.
\end{equation}
Each grid node $g_{xy}$ represents a 1 unit $\times$ 1 unit area (the actual size of 1 unit is defined by the need of different missions), x, y represents the position of the grid in a Cartesian coordinate system. We assume that each cell $g$ in the airspace has a safety score $S(g)$, where $S(g) \in [0, 1]$ and the associated risk score $R(g) = 1 - S(g)$ depending on the situation in that area. The position of the flight agent $p_a$ can only be located at a grid node. Therefore
\begin{equation}
    p_a = (x, y) \quad where: x \in \{1, 2, ...,m\}, y \in \{1, 2, ...,m\}
\end{equation}
means the flight agent is currently on the grid node $g_{xy}$.
Given graph G and the set of edges E, the CSP problem is formulated as the following:
\begin{align}
& \min \sum_{u, v \in N} c_{u v} z_{u v} \label{eq:4.1}\\
& \text { s.t. } \nonumber\\
& \sum_{\{v \mid u, v \in G\}} z_{u v}-\sum_{\{v \mid u, v \in G\}} z_{v u}= \begin{cases}1 & \text { for } u=g^s \\
-1 & \text { for } u=g^t \qquad \forall u \in G \\
0 & \text { otherwise }\end{cases} \label{eq:4.2}\\
& \prod_{z_{u v} = 1} S(v) \geq \epsilon \label{eq:4.3}\\
& \sum_{u, v \in G_s} z_{u v} \leq|G_s|-1 \text { for all } G_s \subseteq G\backslash g^s \label{eq:4.4}
\end{align}
where $c_{u v}$ is the cost of moving from grid $u$ to grid $v$, $z_{u v}$ is a binary variable determining if the path passes $e_{u v}$,  $z_{u v}=1$ if the edge is being traversed, otherwise $z_{u v}=0$.
The objective~(\ref{eq:4.1}) is to minimize the cost along the path subject to three constraints. The first constraint~(\ref{eq:4.2}) enforces the number of edges leading towards a grid is equal to the number of edges leaving that grid. 
The second constraint~(\ref{eq:4.3}) is a safety constraint where the accumulated safety (assuming each step is independent) in the path must be larger than the minimal safety threshold $\epsilon$. The third constraint~(\ref{eq:4.4}) eliminates the sub-tour by limiting the maximum number of edges in the path to the total nodes. $G_s$ is a subset of $\mathrm{G}$ that includes all nodes except for the starting node.

In this paper, we assume the graph is an undirected graph, which means if the agent can move from grid A to grid B, then the agent can move from grid B to grid A. We also assume the agent can only move straight up (y + 1), down (y - 1), left (x - 1), and right (x + 1) for each step. The cost of moving from one grid to an adjacent one is considered to be uniform:
\begin{equation}
c_{u v}=1 \quad where: u \in \mathbf{Adj}(v), v \in \mathbf{Adj}(u), \forall u, v \in G.
\end{equation}

A valid flight plan $T_k$ for flight agent $f_k$ is defined as follows: 

\begin{equation}
    T_k = \{g^{(0)},g^{(1)},..., g^{i},g^{i+1}, ..., g^{(k)}\} \quad \text{where}: \prod_{i=0}^{k}S(g^{(i)})) > \epsilon
\end{equation}
where $g^{(0)}$ is the current position of the agent, and $g^{(1)}$ is the start of the planned trajectory, $g^{(k)}$ is the destination, and any successive $g^{i}$ and $g^{i + 1}$ must be adjacent on the graph G, also $\epsilon$ is the minimal safety  threshold. By assuming each step is independent, the product of the safety score of all the grids in the trajectory must be larger than the minimal safety threshold.
Fig.~\ref{fig:risk} shows an example of an 11*11 risk map, where grayscale represents risk value. In the example map, the start node is the node whose left-bottom corner is located at (2, 8), the initial safety score is 1, the safety constraint threshold is 0.9, and the destination node is the node whose left-bottom is located at (6, 6). The shortest path, which is framed by the red in the figure, apparently goes around the grid with the darker shade to meet the risk constraints. 

\section{Background}\label{sec:literature}
\subsection{A* algorithm and heuristic}
The traditional A* was first proposed in \cite{hart1968formal}. A* is an informed search algorithm, or a best-first search, meaning that it is formulated in terms of weighted graphs: starting from a specific node of a graph, it aims to find a path with the smallest cost to the given goal node.
To aid its performance, A* uses a heuristic function to estimate the cost to reach the goal from a given node. This heuristic adds an additional estimated cost to the path cost from the start node, providing a sort of `guess' of the total path cost.

The A* algorithm makes sure that the path it extends has the smallest possible sum of its past path cost and the estimated cost to the goal. This sum is often denoted as 
\begin{equation}
    f(n) = g(n) + h(n)
\end{equation}
where $f(n)$ is the total estimated cost of the path through the node n, $g(n)$ is the cost from the start node to the node n, $h(n)$ is the heuristic estimated cost from n to the goal.
To guarantee finding the optimal path, the heuristic needs to be admissible. Being admissible means the heuristic value is never overestimating the distance to the goal. If the heuristic is monotone, A* does not need to re-open closed nodes and can be more efficient. Being monotone means the heuristic value follows triangle inequality, which means its estimate is always less than or equal to the estimated distance from any neighboring vertex to the goal, plus the cost of reaching that neighbor.

\subsection{Traditional way to accelerate A* algorithm}
Manhattan and Euclidean heuristics are very useful admissible heuristics because the distance between the two nodes can not be shorter than the Euclidean distance (Manhattan in the case diagonal movement is not allowed).
A* Triangle Inequality Landmarks(ATL) method \cite{goldberg2005computing} is a heuristic for A* relied on pre-calculating ``landmarks'' on a graph that could then be used with the triangle inequality to find minimum distances quickly. The triangle inequality states that the sum of the lengths of any two sides of a triangle must exceed the length of the third side. However, the ATL principle does not hold in the context of the CSP problem. For example, consider using grid A as a landmark. we identify the shortest path from grid A to grid B,  denoted as $p_{ab}$ and from grid A to grid C denoted as $p_{ac}$, where the length of $p_{ab}$ is greater than that of $p_{ac}$: $len(p_{ab}) > len(p_{ac})$. If there is no resource constraint, the length of the shortest distance between grid B and grid C must be longer than $len(p_{ab}) - len(p_{ac})$, otherwise, $p_{ab}$ would not be the shortest path between grid A and B because there exists a shorter path $p_{acb}$.  However, with resource constraints, this inequality does not necessarily hold, as the agent might be unable to travel through the path $p_{acb}$. 
The Jump Point Search(JPS) algorithm, introduced by \cite{harabor2011online}, is one such way of making pathfinding on a rectangular grid more efficient. JPS only stops at nodes that are directly reachable and might affect the optimal path. However, in the CSP problem, every node affects the optimal path because every node may consume resources.

\section{Method}\label{sec: method}
\subsection{Additional safe dimension A* (ASD A*)}
Given a start grid and a goal grid, the traditional A* can find the optimal path while expanding fewer grids. However, the traditional A* algorithm can not record the safety consumption. Therefore, an additional dimension needs to be considered. In this paper, we introduced the additional safety dimension A* (ASD A*) to search the constrained shortest path in the risk map. Each node has an extra safety dimension besides the position. The extra safety of each node will be equal to the product of the safety of the previous node in the path and the safety score of this node. Because the safety score of all the nodes is smaller than 1, the safety of the node monotonically decreases along the path. Like traditional A*, the ASD A* algorithm first initials a priority queue and adds the start node to it. The node in the priority queue is sorted by f-value. We explore the node with the lowest f-value in each iteration. Unlike the traditional A*, each node now has three dimensions, x, y, and safety. When A* discovers a new node that was on an old grid explored before, A* will compare the f-value of two nodes on the same grid and only keep the one with a lower f-value. In contrast, ASD A* will compare both the f value and safety of the new nodes to every existing node. If the new node has a lower f-value and higher safety than the existing node, the new node will be kept and the old node will be removed. If the new node has a higher f-value and higher safety, both nodes will be kept. If the new node has a higher f-value and lower safety, then the new node will not be added to the priority queue. The algorithm keeps exploring the node until the destination is found. This algorithm can guarantee finding the shortest path if the h-value is admissible. If there exists a path shorter than the founded path, the node in this path will be explored before the final node.   

\begin{figure}
    \centering
    \includegraphics[width=0.5\linewidth]{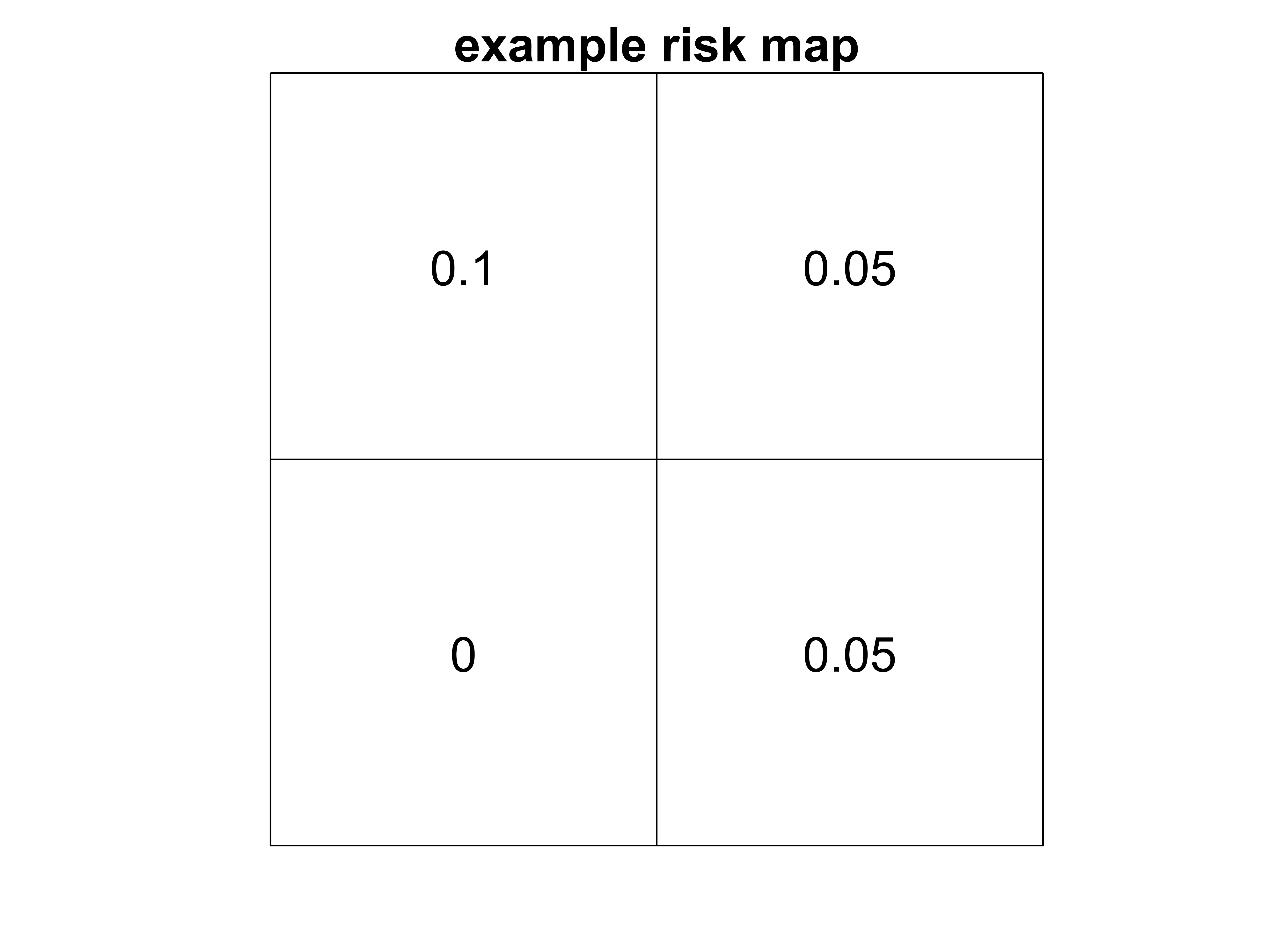}
    \caption{Example risk map}
    \label{fig:riskmapexample}
\end{figure}

The following is an example search for the constrained shortest path using the ASD A* algorithm with the Manhattan heuristic. The example risk map is shown in Fig~\ref{fig:riskmapexample}, the start grid $g_{00}$ is the left-bottom grid with $R(g_{00}) = 0$, grid $g_{01}$ is the left-top node with $R(g_{01}) = 0.1$, grid $g_{10}$ is the right-bottom node with $R(g_{10}) = 0.05$, and the destination grid $g_{11}$ is the right-top node with $R(g_{11}) = 0.05$. Let the safety constraint threshold be 0.9. In the grid world, there are four direct neighbors, the upper grid, the down grid, the right grid, and the left grid for all the grids except those near the boundary. 
The node of the upper grid is added to the priority queue first, followed by the nodes of the right and left grid, and the node of the down grid is added last.

\begin{figure}
    \centering
    \includegraphics[width=0.5\linewidth]{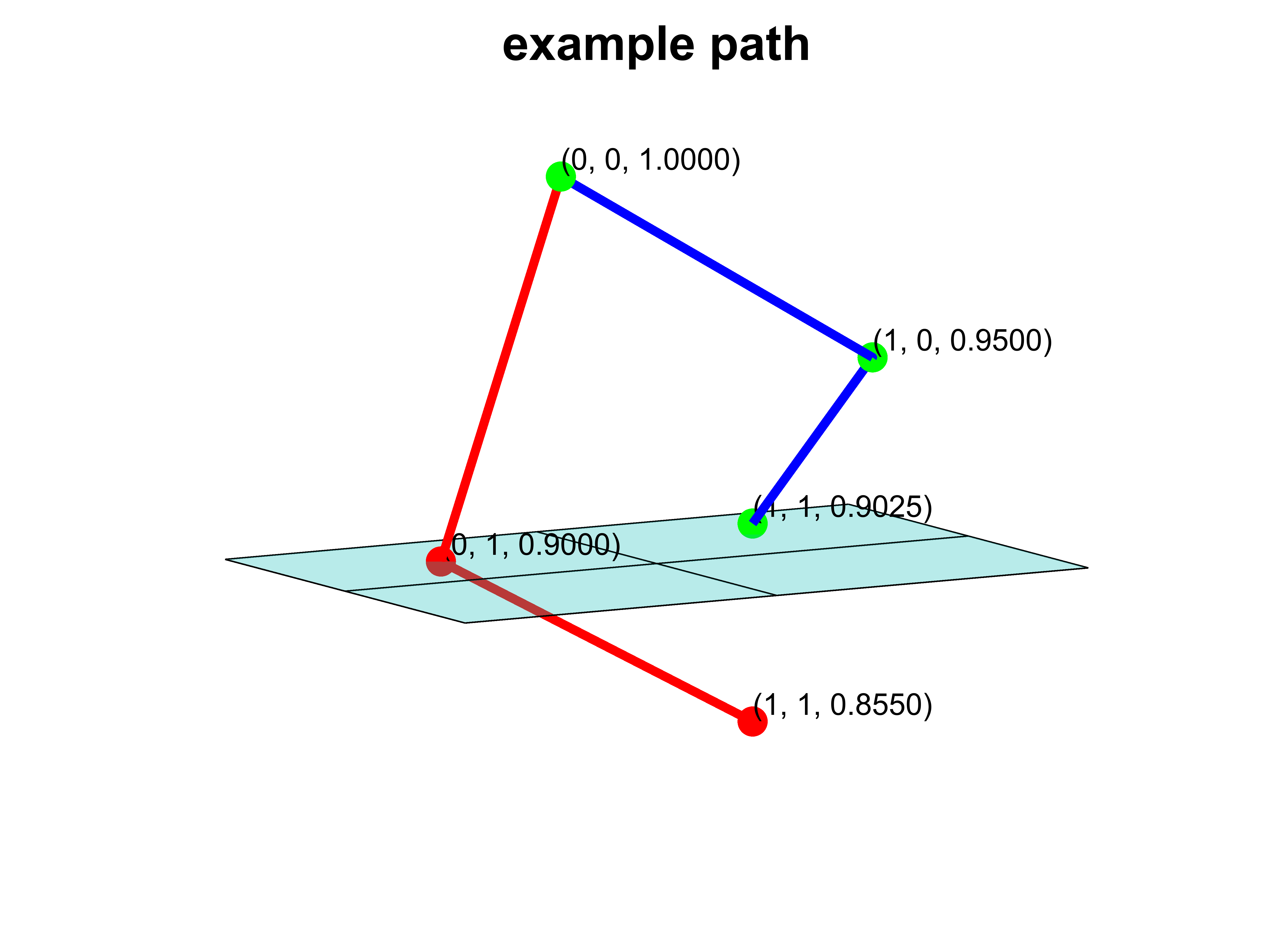}
    \caption{Example path found in risk map}
    \label{fig:riskmapexamplepath}
\end{figure}

The ASD A* algorithm first initials a priority queue and adds the start node to it. Because $R(g_{00}) = 0$, the node (0, 0, 1) is added to the priority queue. Then we explore the node (0, 0, 1) since it is the only node in the priority queue. As the agent can move straight up and straight right, we found $g_{01}$ and $g_{10}$. Because $R(g_{01}) = 0.1$, the safety of the new node is equal to $1 * (1 - 0.1) = 0.9$, then the new node (0, 1, 0.9) is added to the priority queue with the f-value equal to 2. Then, the new node (1, 0, 0.95) is also added to the priority queue with the f-value equal to 2. Then the algorithm explores the node (0, 1, 0.9), and the two reachable neighbors are (0, 0, 0.9) and (1, 1, 0.855). Because node (0, 0, 0.9) has a larger f-value and lower safety than the explored (0, 0, 1), this node will be ignored. For (1, 1, 0.855), although it reaches the destination, the safety is lower than the safety constraint, this node is also invalid. Therefore, the path $T^1 = \{g_{00}, g_{01}, g_{11}\}$ shown as the red edges in Fig~\ref{fig:riskmapexamplepath} is not feasible. Then we explore the node (1, 0, 0.95), and we can find (0, 0, 0.95) and (1, 1, 0.9025). The node (0, 0, 0.95) will be ignored. The node (1, 1, 0.9025) reaches the destination while it meets the safety constraint. Therefore the feasible path $T^2 = \{g_{00}, g_{10}, g_{11}\}$ as the blue edges in Fig~\ref{fig:riskmapexamplepath} is found.


\subsection{Expert heuristic generating}
ASD A* algorithm with the Manhattan heuristic is very time-consuming because the new dimension scales up the number of total exploreable nodes. Therefore, designing a better heuristic is very important. We first create the expert heuristic that can lead ASD A* to find the optimal path as fast as possible without factoring in time-consuming. For each start-destination pair, we start with the Manhattan heuristic. Then we find the optimal path and lower the h-value of all the nodes in the optimal solution path so that ASD A* will explore them first. Finally, we find the nodes that the agent can not arrive at the destination node if the agent passes them. We mark them as infeasible so the ASD A* does not need to explore them. It is very time-consuming to generate the expert heuristic because we are brute-force every possible solution.

\subsection{Transformer-based Neural networks}
\begin{figure}
    \centering
    \includegraphics[width=0.75\linewidth]{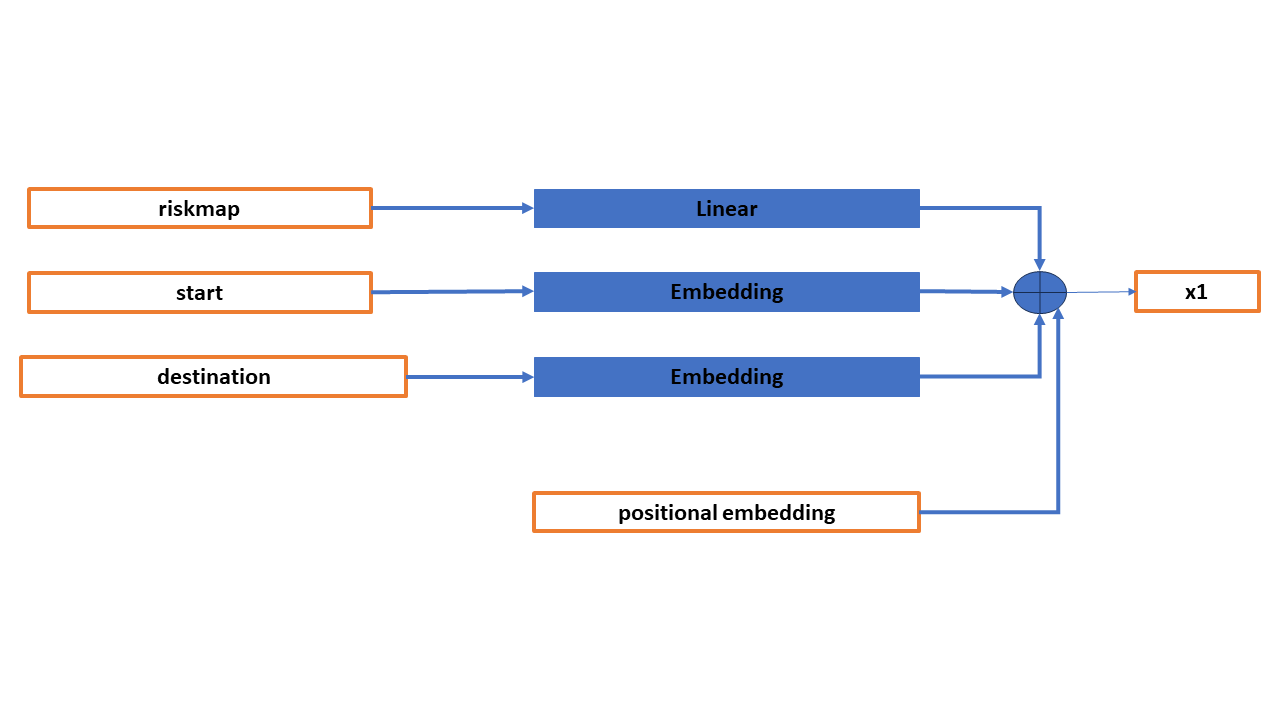}
    \caption{Input layer}
    \label{fig:input}
\end{figure}
\begin{figure}
    \centering
    \includegraphics[width=0.75\linewidth]{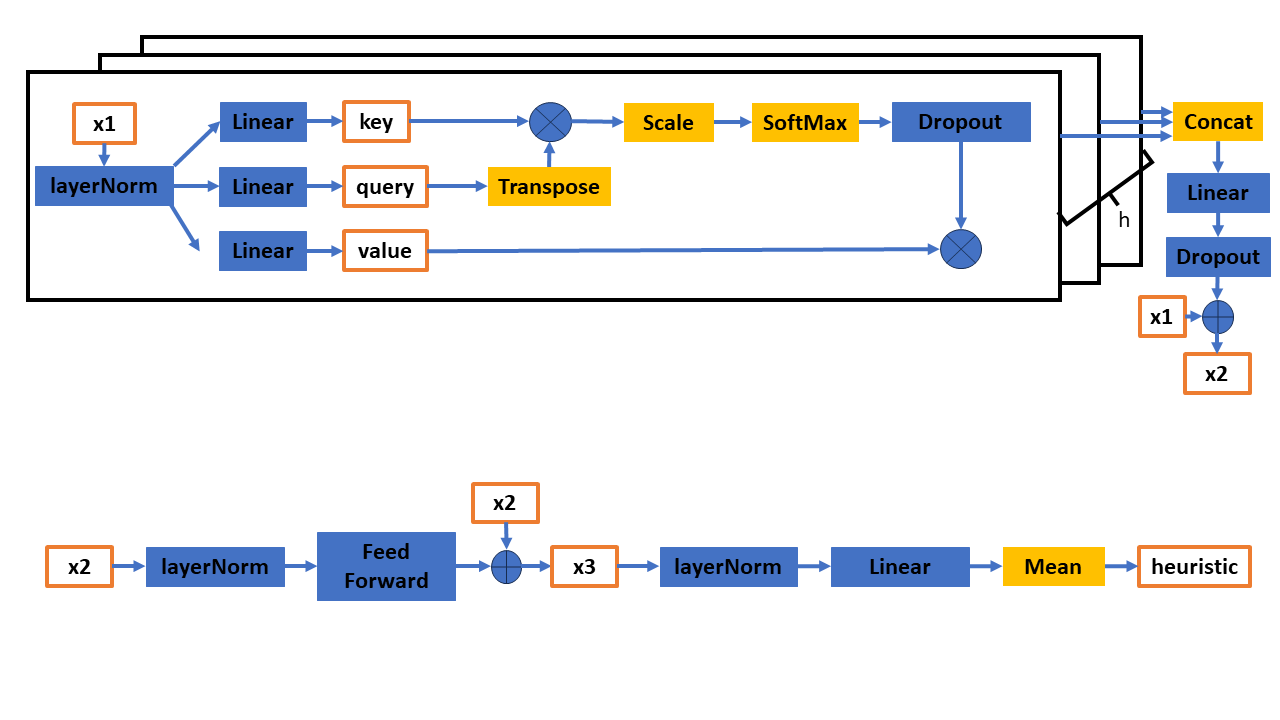}
    \caption{Backbone and output layer}
    \label{fig:backbone}
\end{figure}

In this paper, a transformer-based neural network will be used to generate the heuristic value given the risk map, the starting grid, and the destination grid. Because a good heuristic highly relies on every grid in the risk map, neural networks must be able to let each grid in the risk map attend to heuristic generating in all other grids. Consequently, the self-attention mechanism, being a fundamental component of the transformer architecture, emerges as the most suitable neural network model for this purpose so far. The input layer of the neural network is shown as Fig~\ref{fig:input}, and the Backbone neural network and output layer are shown as Fig~\ref{fig:backbone}.

In the figures, the orange outline boxes are the vectors and
the blue rectangles are neural network layers. The Linear means linear layer, which applies a linear transformation to its input data. Mathematically, the linear layer can be represented by the following equation:
\begin{equation}
Linear(x) =x W^T+b,
\end{equation}
where $W$ is the weight matrix, $b$ is the bias vector, and $x$ is the input.
Embedding means embedding layer, which is a special layer used to transform discrete input data into dense vectors. It is simply a trainable table, that outputs dense vectors based on the discrete input data. The Dropout means dropout technique~\cite{hinton2012improving}. Dropout randomly deactivates a subset of neurons in the layer for an iteration to prevent the network from overfitting. The LayerNorm means layer normalization~\cite{ba2016layer}. Layer normalization normalizes the inputs across the features, which helps to stabilize training. The Feed Forward is position-wise feed-forward networks~\cite{vaswani2017attention}, consists of two linear transformations with a ReLU activation in between, and can be represented by the following equation:
\begin{equation}
\operatorname{FFN}(x)=\max \left(0, x W_1+b_1\right) W_2+b_2,
\end{equation}
where $W_1$ and $W_2$ are the weight matrix of linear layer, $b_1$ and $b_2$ are the bias vector, $x$ is the input.

The yellow rectangles are mathematical functions. The Transpose simply transposes the input matrix. The Scale means scale function which divides the input by the square root of the dimension of the key vector. This is an important process in the self-attention mechanism to decrease the value of the dot product of query and key vectors. The Softmax means the softmax function to normalize the input. The Concat means concatenation process which combines all the output into single vectors end-to-end. The Mean means mean function which computes the average value along a specified dimension of a vector. 

The input of the neural network includes the risk map vector, start index, destination index, and positional embedding. The risk map vector is a one-dimensional vector obtained by flattening a 2D risk map. The start and destination are indexes indicating the positions of the starting and destination grids within this flattened vector. For example, if the risk map's shape is $2\times2$, the risk map vector will be $4\times1$, the grid (0,0) will be integer 0, and the grid (1,1) will be integer 3. Positional embedding is simply embedded x-term arithmetic sequence, where x is equal to the length of the risk map vector $d_r$. Positional embedding is crucial for the model to understand the sequence and structure of the input data.

In the input layer, the risk map is processed through a linear layer, while the start and destination index are passed to an embedding table. This ensures that all four inputs attain the same dimensional shape. These inputs are then summed to form a unified vector, which we will refer to as $x1$. 
Then, the processed input $x1$ will be passed to the self-attention mechanism.
The $x1$ will be sent to three different linear layers to get the key, the query, and the value. Then we compute the dot products of the query with the key, divide each by the dimension of the key, and apply a softmax function. Then the output is obtained by taking the dot product of the softmax function's output with the value matrix. In practice, the attention function is computed simultaneously for a group of queries, collectively organized into a matrix $Q$. Similarly, the keys and values are aggregated into matrices $K$ and $V$ respectively. 
The self-attention mechanism can be written in the following equation:
\begin{equation}
\operatorname{Attention}(Q, K, V)=\operatorname{softmax}\left(\frac{Q K^T}{\sqrt{d_k}}\right) V,
\end{equation}
where $d_k$ is the dimension of the key. 
In the transformer, the self-attention mechanism is executed in parallel $h$ times, each time with the same input $x1$ but processed through different linear layers. Then, all the output of the self-attention mechanism will be concatenated together and passed to a linear layer. The output is then added to the initial input $x1$, resulting in the final output $x2$. This procedure is known as a residual connection~\cite{he2016deep}.

In the output layer, the vector x2x2 undergoes several steps: First, it is passed through LayerNorm, followed by a Feed Forward. After that, a residual connection is applied, followed by another LayerNorm operation. Finally, the data passes through a Linear layer. The output vector of the final Linear layer is a 2D vector of size $d_r\times d_r$, and applying the mean function along one dimension results in a final output of size $d_r \times 1$. 

To train the neural network, a custom loss function is developed. This function comprises two parts: the first is the Mean Squared Error (MSE), which ensures the network's output closely aligns with that of the expert. The second component is a penalty for overestimation, implemented to maintain the admissibility of the output. The custom loss (CL) function is shown as the following:
\begin{equation}
CL\left(y_{\text {output }}, y_{\text {expert }}\right)=\operatorname{MSE}\left(y_{\text {output }}, y_{\text {expert }}\right)+\alpha \cdot \frac{1}{n} \sum_{i=1}^n \max \left(y_{\text {output }, i}-y_{\text {expert }, i}+\beta, 0\right),
\end{equation}
where $\alpha$ is a scaling factor for the penalty term and $\beta$ is a small constant added to the difference between the prediction and the target in the penalty term.

\section{Result}\label{sec: results}

In this paper, we evaluate the proposed method in two types of $16 \times 16$ grid world datasets (random map and wind flow map). For each risk map, only the suitable start-destination pairs can generate sample cases. Two requirements make a start-destination pair suitable. Firstly, the Manhattan distance between the start grid and destination grid must exceed 10. Secondly, there is a feasible path between the two grids. 

For the random map, the training dataset has 471 maps and 4896699 cases, and the test dataset has 52 maps and 528952 cases. All the test cases have the same minimal safety threshold $\epsilon = 0.9$.  For the wind flow map, the training dataset has 995 maps and 9339265 cases, and the test dataset has 84 maps and 802934 cases. All the test cases have the same minimal safety threshold $\epsilon = 0.8$. We will assess the effectiveness of the learned heuristic using four distinct criteria: I) we'll examine the average number of nodes explored per path found, for which is a crucial measure to evaluate the efficiency of the A* algorithm. A lower count of explored nodes indicates better efficiency; II) we'll consider the total time taken per path found, including both the time for generating the heuristic and the time for the A* algorithm's search process; III) The third metric is the average length of the path found. This metric can determine if the founded path is optimal; IV) The fourth will be the win ratio of faster test cases. If the learned heuristic costs more time or has a longer path, the Manhattan heuristic wins, otherwise, the learned heuristic wins. 

The A* algorithm operates on an Intel i9-12900KF CPU, while the heuristic generation process is executed on an Nvidia RTX3090 GPU. For the random dataset, the model underwent 50 training iterations, taking approximately 350 hours on a single Nvidia RTX3090 GPU. In the case of the wind flow dataset, the training involved 10 iterations and required about 250 hours, also on a single Nvidia RTX3090 GPU.

\subsection{Random risk map}
\begin{figure}[!ht]
	\begin{center}
		\subfigure[risk map example 1]{
			\label{fig:fulltree}
			\includegraphics[width=0.48\textwidth]{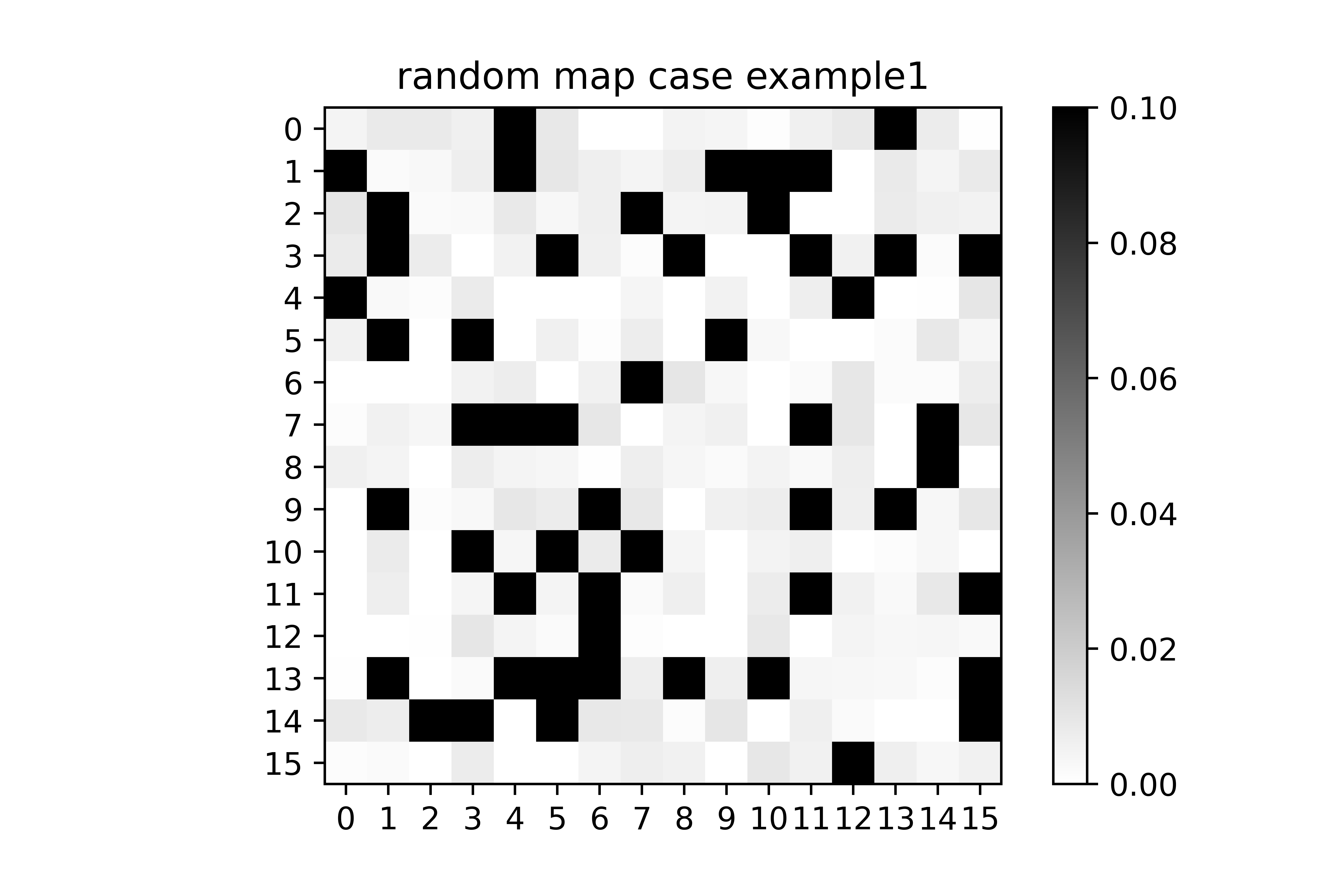}
		}
    	\subfigure[risk map example 2]{
			\label{fig:fulltree}
			\includegraphics[width=0.48\textwidth]{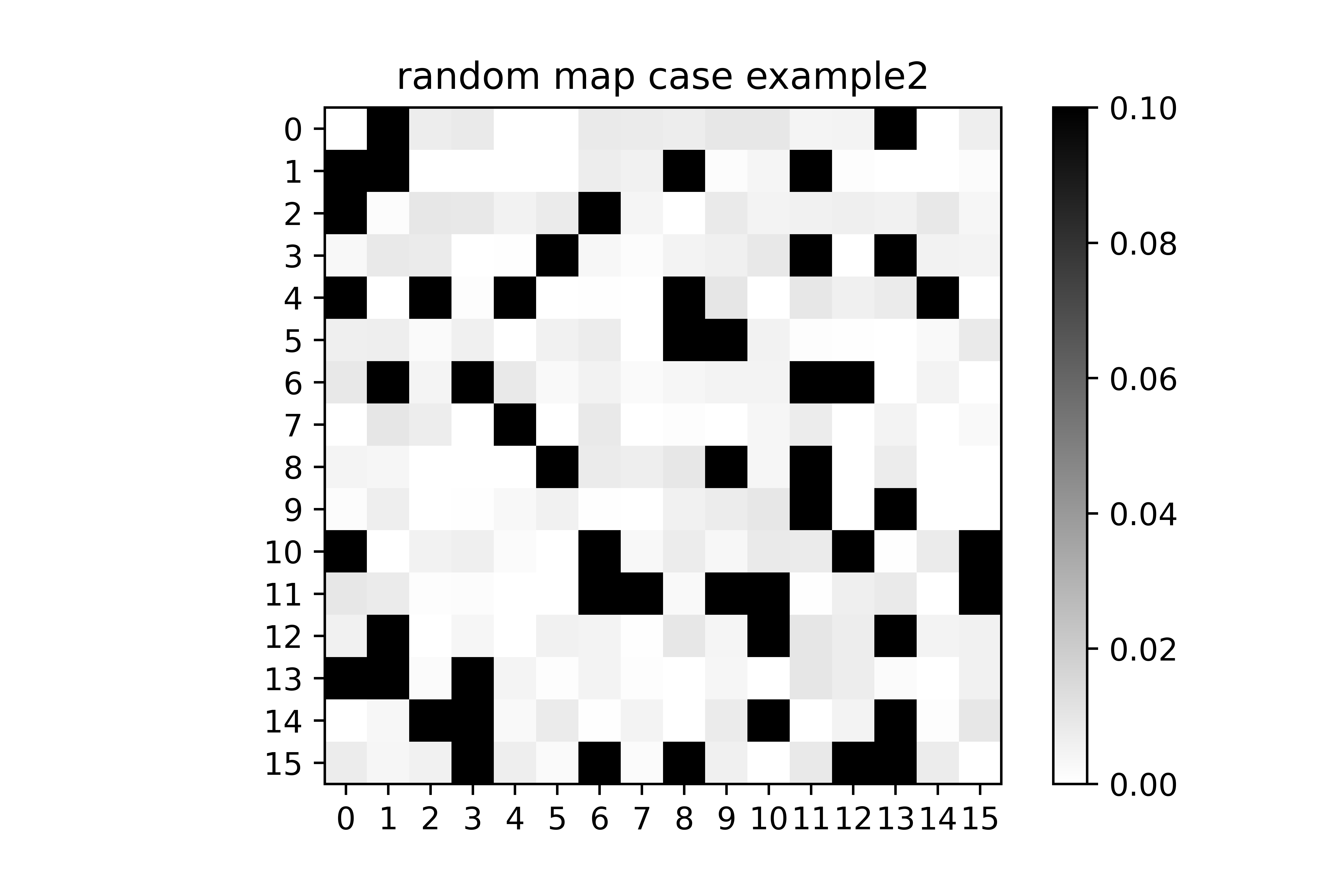}
		}
		\subfigure[example 1 with Manhattan]{
			\label{fig:fullmap}
			\includegraphics[width=0.48\textwidth]{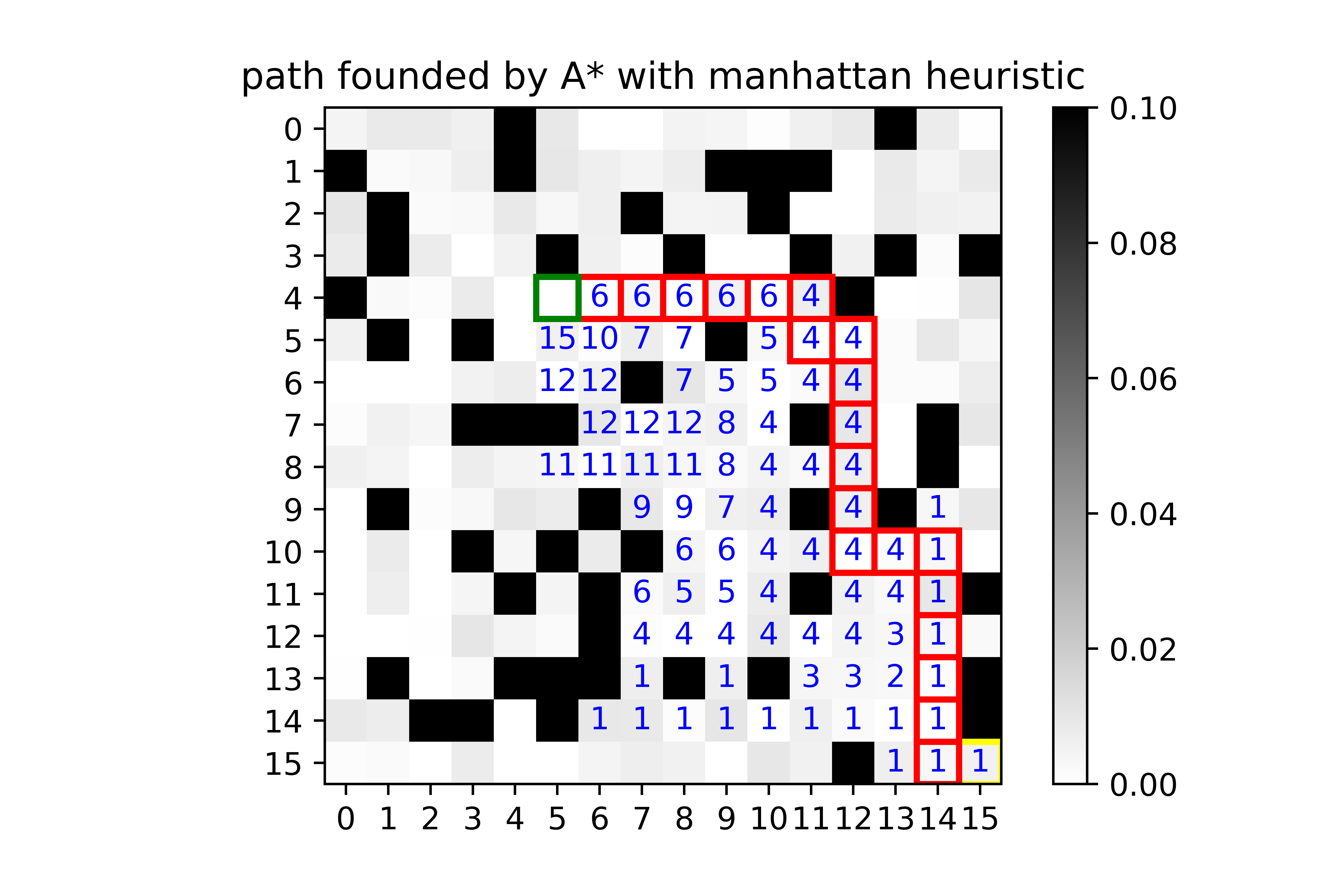}
		}
  		\subfigure[example 2 with Manhattan]{
			\label{fig:fullmap}
			\includegraphics[width=0.48\textwidth]{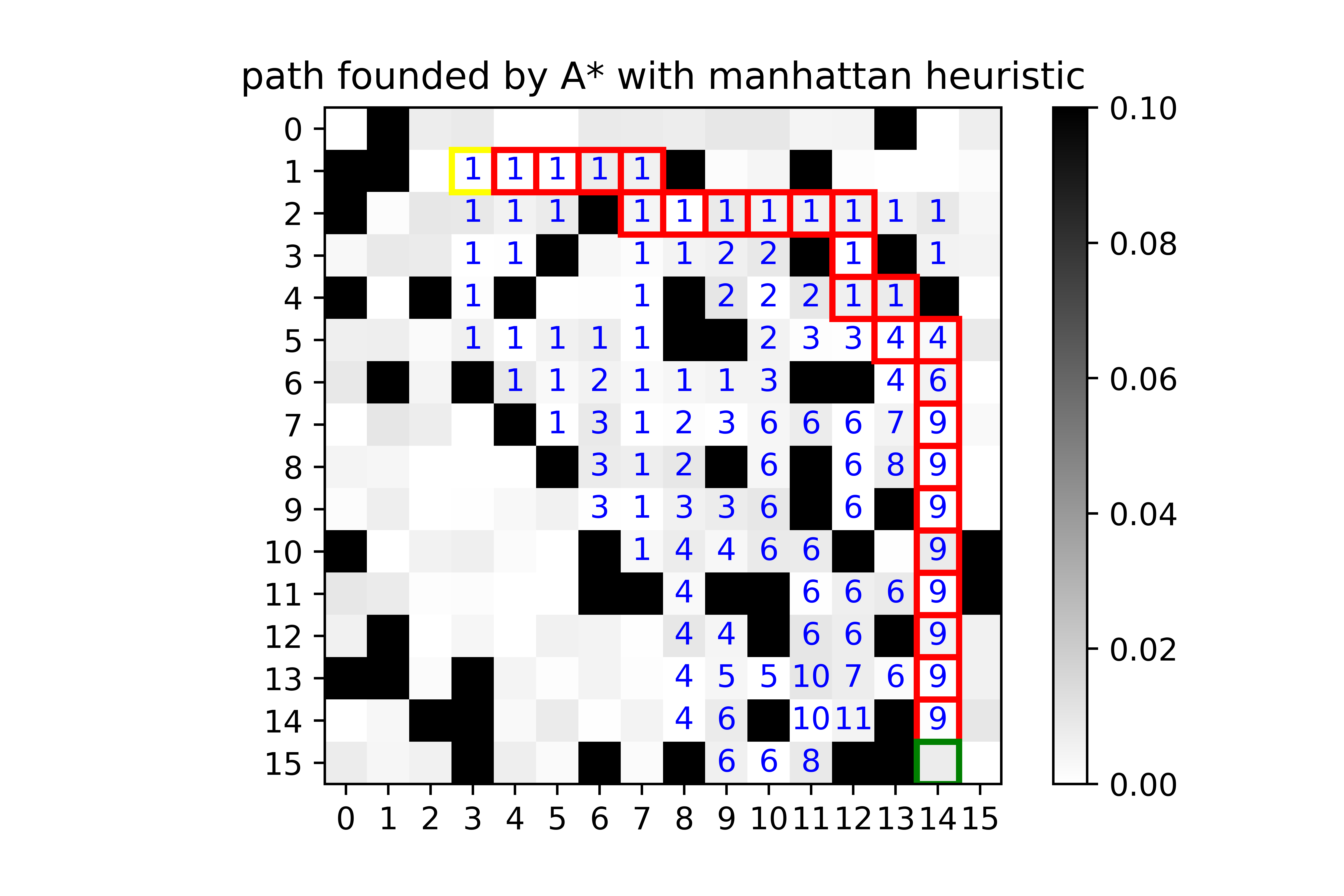}
		}
		\subfigure[example 1 with learned heuristic]{
			\label{fig:partialtree}
			\includegraphics[width=0.48\textwidth]{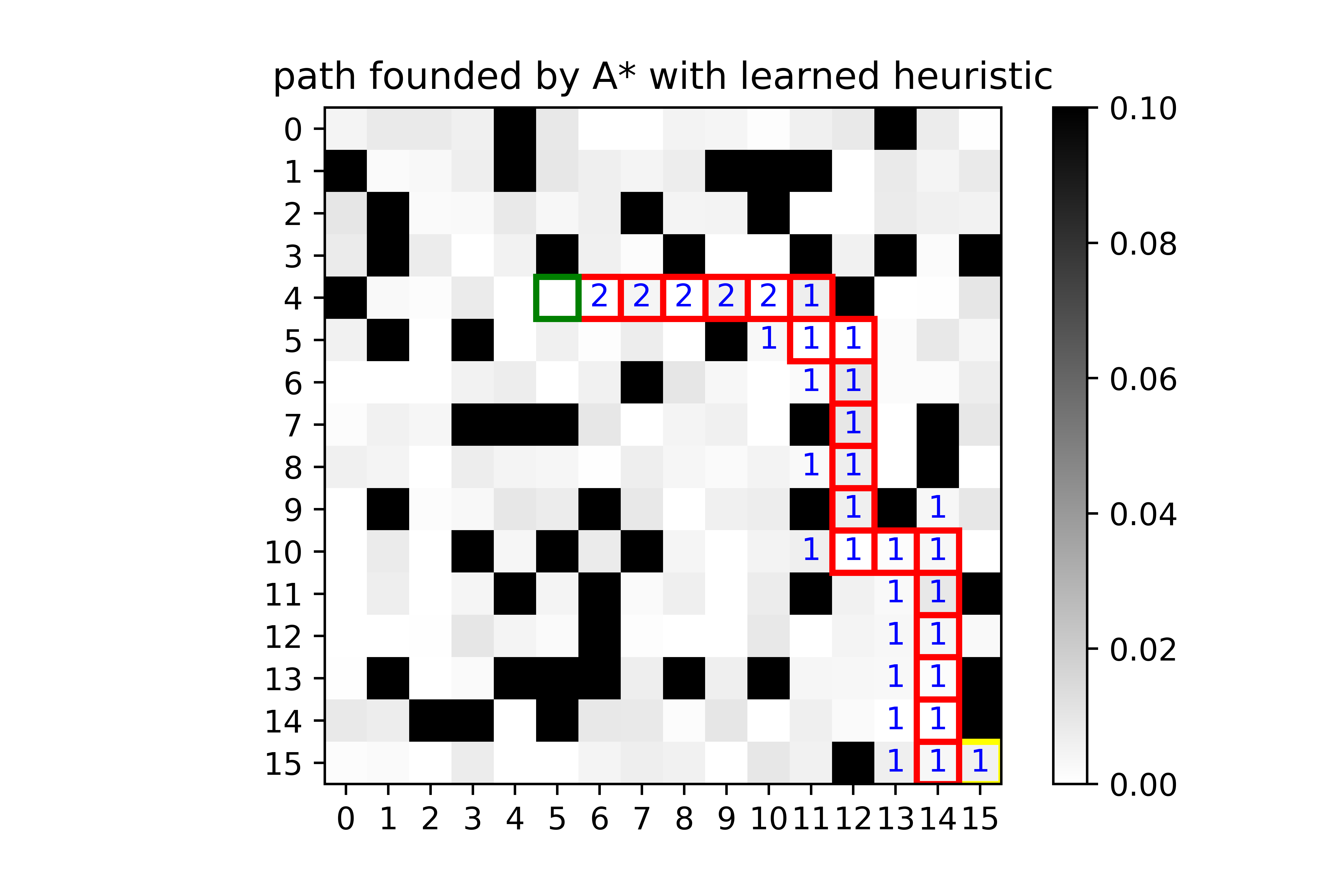}
		}
		\subfigure[example 2 with learned heuristic]{
			\label{fig:partialtree}
			\includegraphics[width=0.48\textwidth]{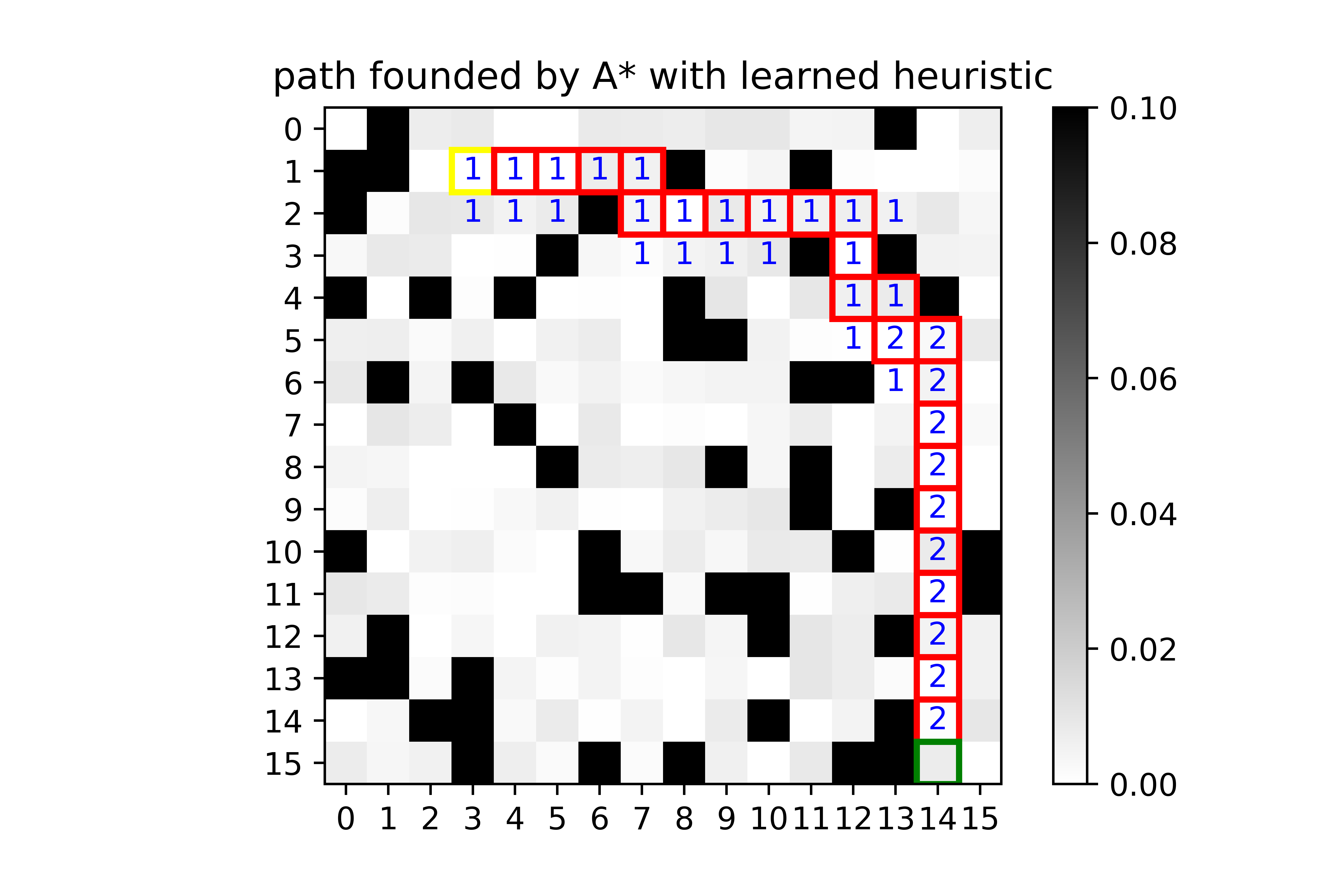}
		}
	\end{center}
	\vspace{-0.3cm}
	\caption{Random risk map examples. The yellow box represents the starting grid, the green box is the destination grid, and the red boxes denote the path. The blue numbers indicate how many times each grid is explored by the ADS A* algorithm.}
	\label{fig:randomexamples}
\end{figure}

 The first type is a random map where 20\% of the grids are entirely safe (safety score $R(g)=1$), allowing the agent to traverse without any safety loss. Another 20\% of grids are completely dangerous (safety score $R(g)=0$), so the agent can not pass them. The remaining 60\% of grids are risky grids, so the agent will lose some safety passing them. The risk of arriving at risky grids is between 0 and 0.02. The Fig.~\ref{fig:randomexamples} presents two examples of a random map, where the yellow box represents the starting grid, the green box is the destination grid, and the red boxes denote the path. The blue numbers indicate how many times each grid is explored by the ADS A* algorithm. These examples demonstrate that the A* algorithm, with both the Manhattan heuristic and the learned heuristic, can find the same optimal path. However, it's noteworthy that the A* algorithm with the learned heuristic explores fewer nodes in comparison.

The results for large-scale testing are presented in Table~\ref{tab: randomresult}.
The analysis shows that, compared to the ASD A* algorithm using the Manhattan heuristic, the ASD A* with the learned heuristic reduces node exploration by 38\% and saves 43\% in time on random maps,  while ensuring the optimal path is found. It is also observed that in 85.4\% of the test cases, the ASD A* with the learned heuristic demonstrates faster performance.

\begin{table}[hbt!]
\caption{\label{tab: randomresult} Result in random risk map}
\centering
\begin{tabular}{c c c c}
\hline
\hline
risk map & result & Manhattan & learned heuristic (our method) \\ \hline
\multirow{5}{*}{total number of test case} & total number of test case & 528952 & 528952 \\
& average nodes explored & 232.2 & 167.1 \\
& average time cost (ms) & 42.34 & 24.13 \\ 
& average path length & 18.81 & 18.81 \\
& faster test case ratio & 14.6\% & 85.4\% \\ \hline
\hline
\end{tabular}
\end{table}

\subsection{Wind flow risk map}

\begin{figure}[!ht]
	\begin{center}
		\subfigure[risk map example 1]{
			\label{fig:fulltree}
			\includegraphics[width=0.48\textwidth]{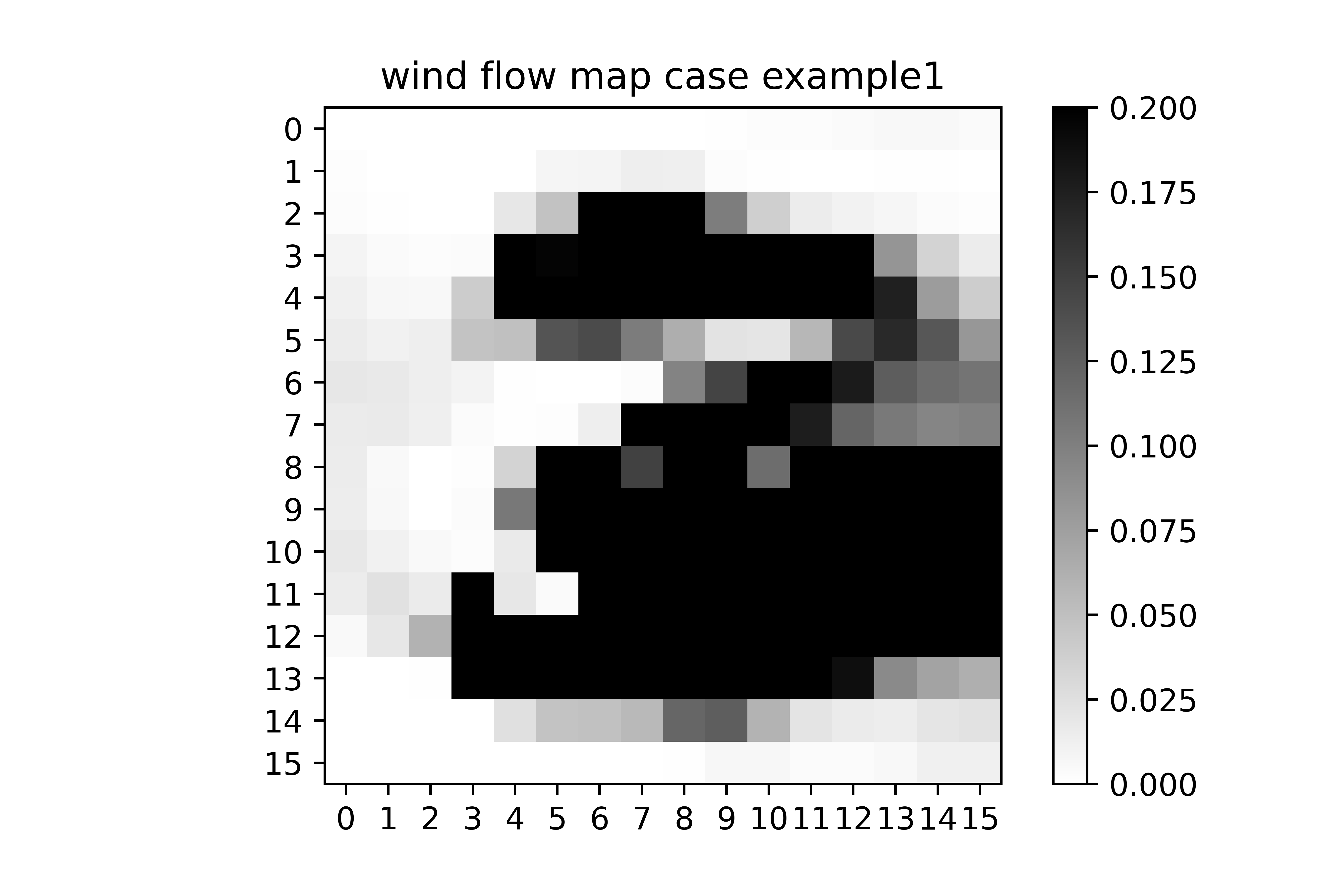}
		}
    	\subfigure[risk map example 2]{
			\label{fig:fulltree}
			\includegraphics[width=0.48\textwidth]{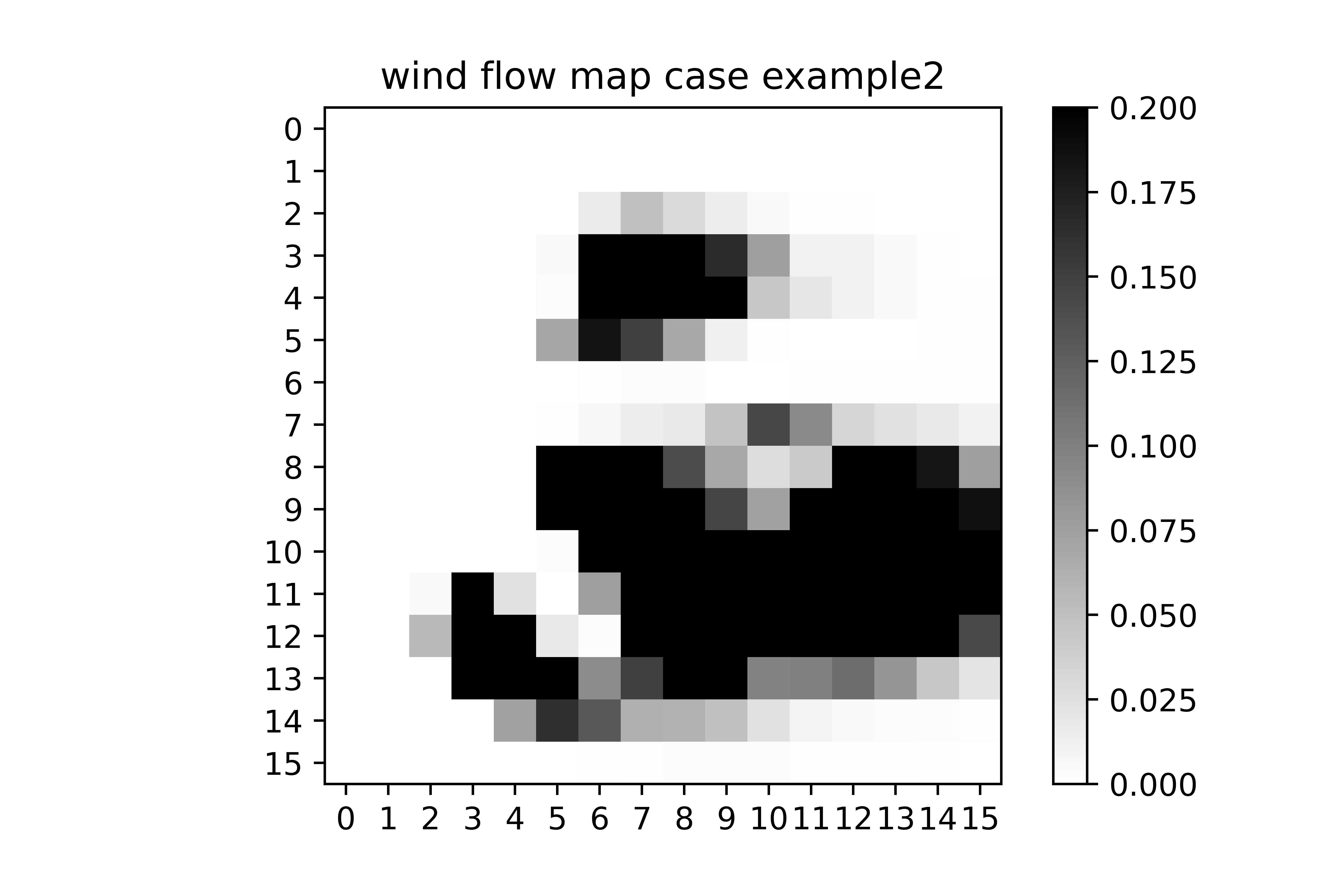}
		}
		\subfigure[example 1 with Manhattan]{
			\label{fig:fullmap}
			\includegraphics[width=0.48\textwidth]{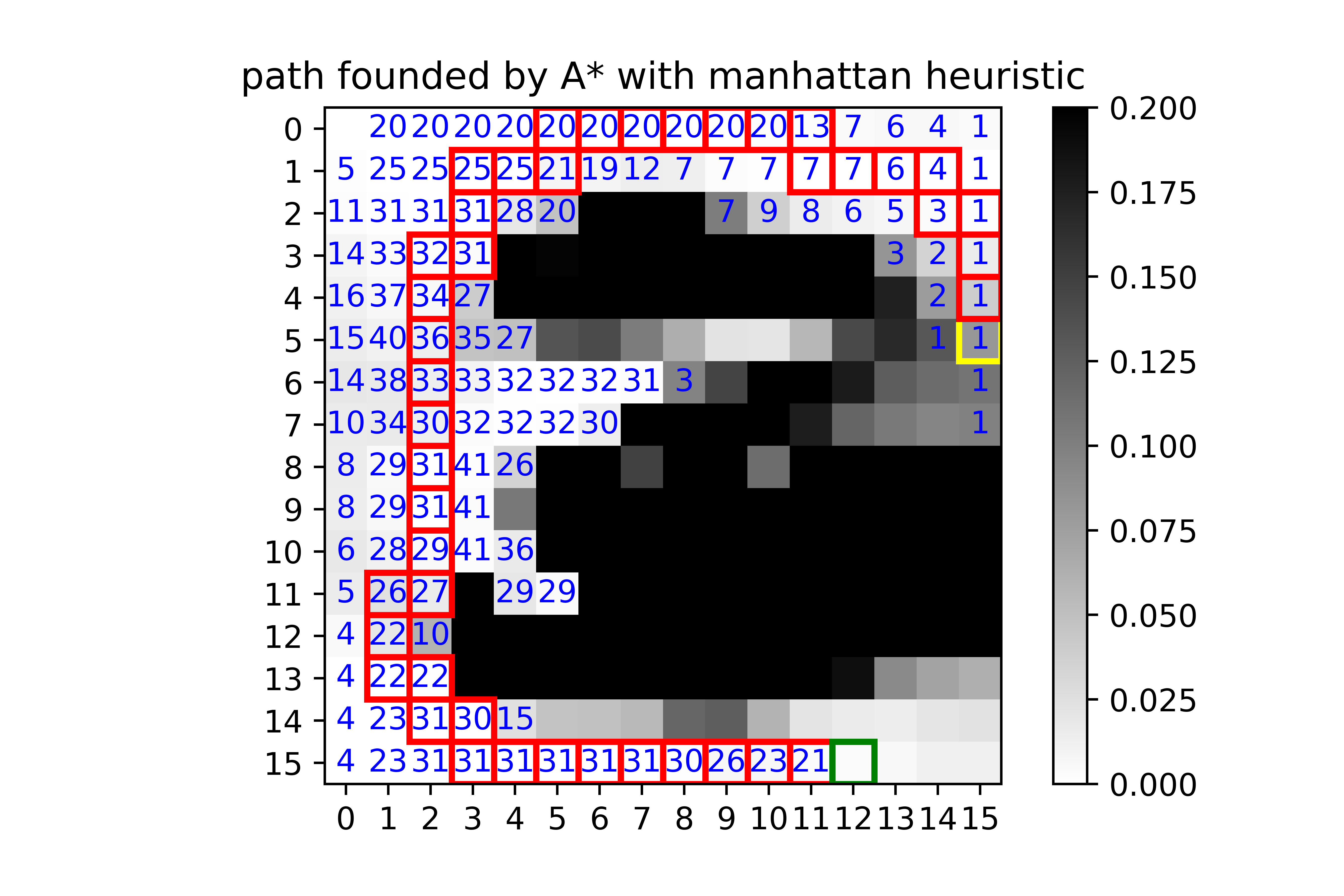}
		}
  		\subfigure[example 2 with Manhattan]{
			\label{fig:fullmap}
			\includegraphics[width=0.48\textwidth]{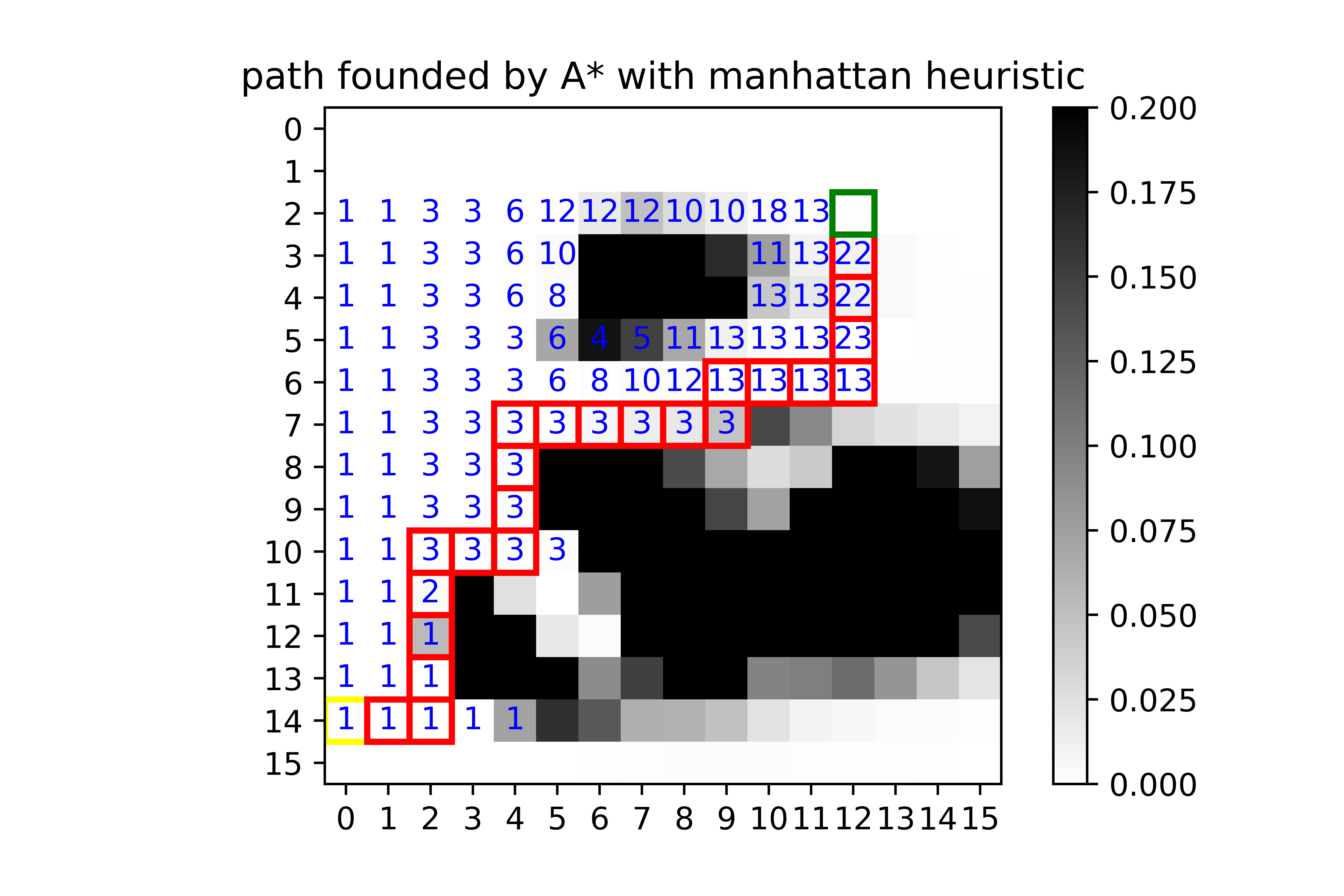}
		}
		\subfigure[example 1 with learned heuristic]{
			\label{fig:partialtree}
			\includegraphics[width=0.48\textwidth]{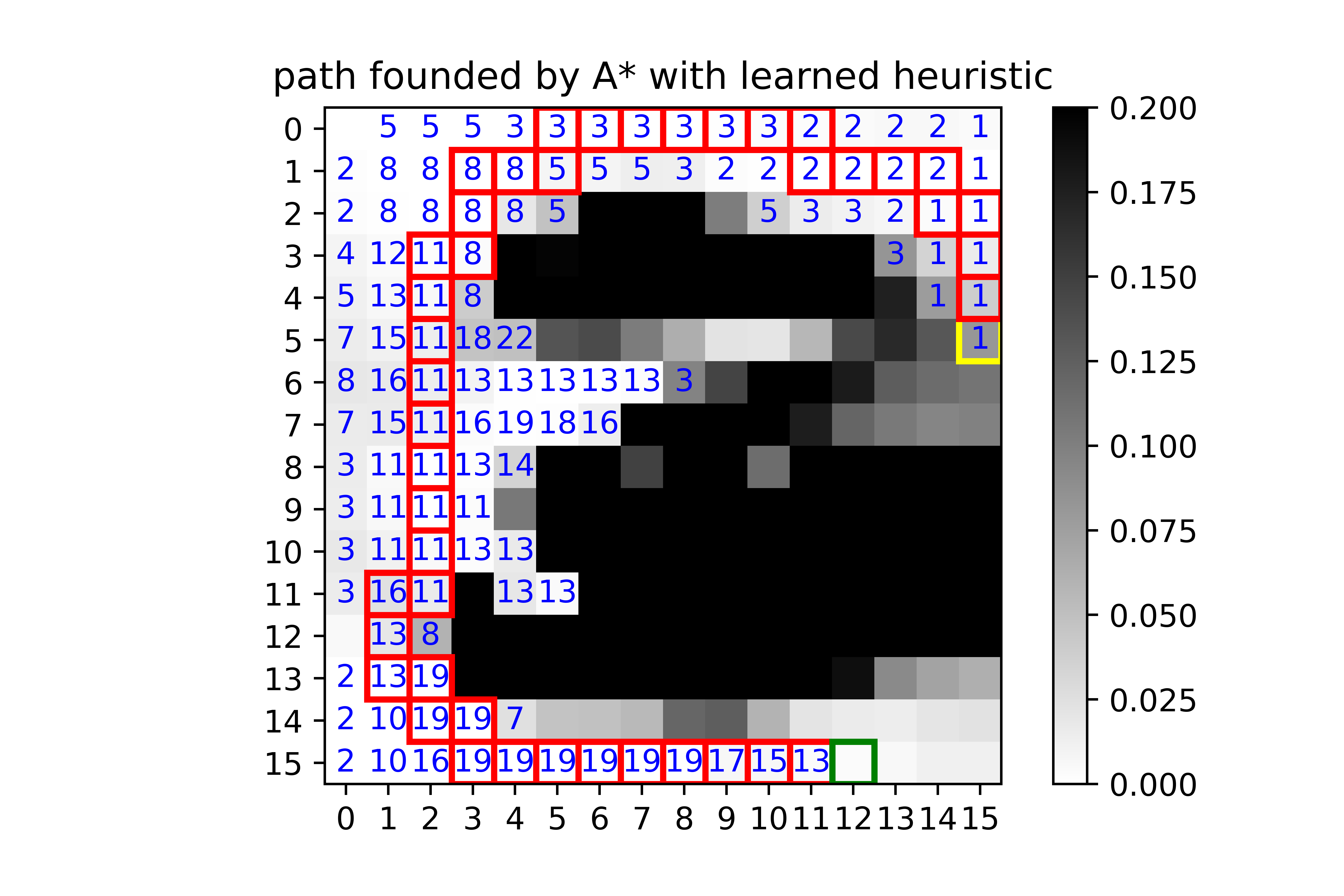}
		}
		\subfigure[example 2 with learned heuristic]{
			\label{fig:partialtree}
			\includegraphics[width=0.48\textwidth]{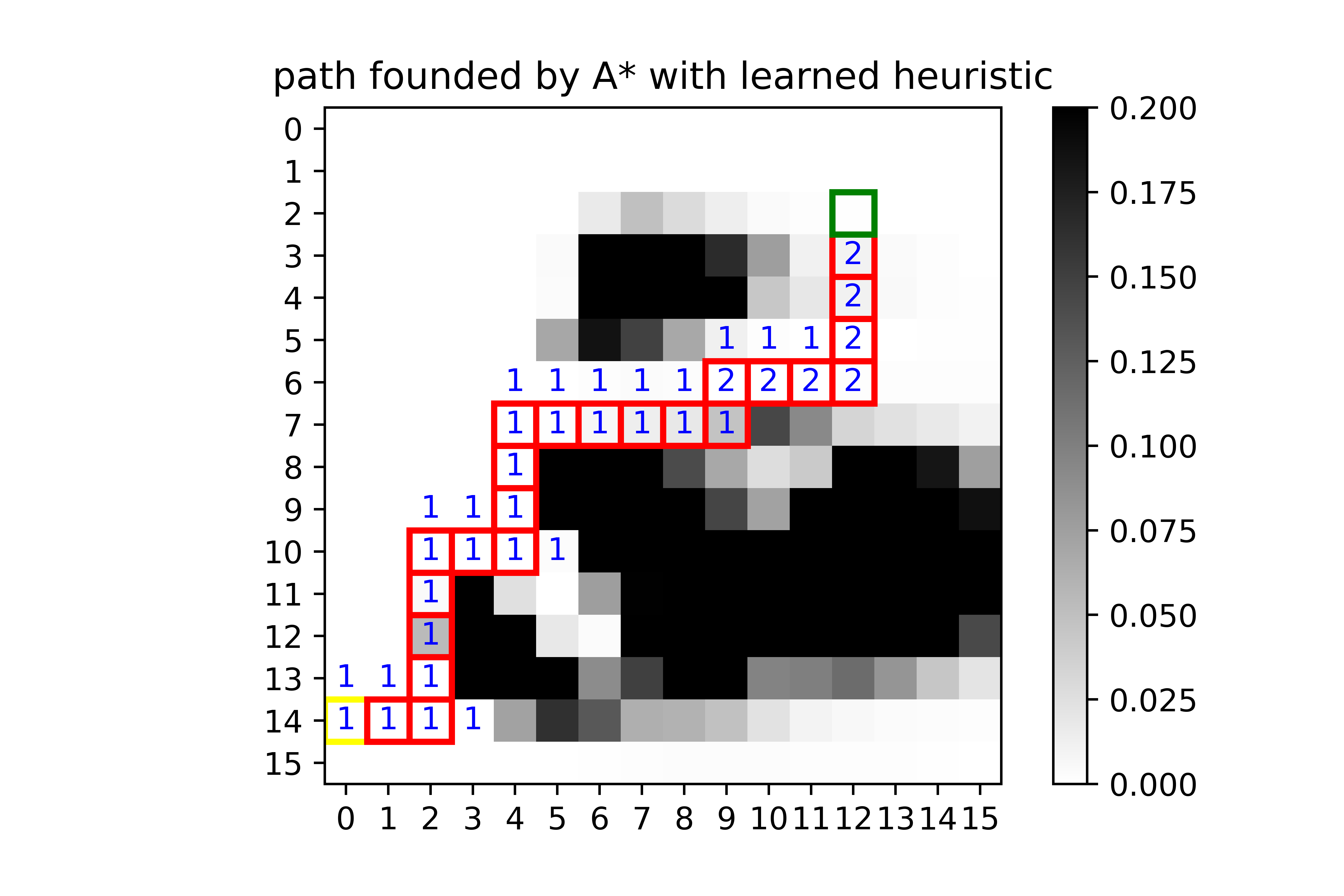}
		}
	\end{center}
	\vspace{-0.3cm}
	\caption{Wind flow risk map examples. The yellow box represents the starting grid, the green box is the destination grid, and the red boxes denote the path. The blue numbers indicate how many times each grid is explored by the ADS A* algorithm. Example 1 has higher height and faster wind, so the risk of most grids are generally higher.}
	\label{fig:windexamples}
\end{figure}

The second type is the wind flow risk map near high buildings generated by a city wind flow simulator. The city wind flow simulator can calculate the risk of each grid for turbulence 
with the position of the buildings, the wind speed, and the assessment height. Fig.~\ref{fig:windexamples} shows the two examples of the wind flow risk map, where the yellow box represents the starting grid, the green box is the destination grid, and the red boxes denote the path. The blue numbers indicate how many times each grid is explored by the ADS A* algorithm. Example 1 has higher height and faster wind, so the risk of most grids are generally higher.

The results are presented in Table~\ref{tab: result}.  Similarly, in wind flow maps, our approach leads to a 52\% reduction in node exploration and a 65\% time saving, again ensuring the optimal path is identified. In 95.6\% of the test cases, the ASD A* with the learned heuristic outperforms Manhattan in terms of speed.

\begin{table}[hbt!]
\caption{\label{tab: result} Result in wind flow map}
\centering
\begin{tabular}{c c c c}
\hline
\hline
risk map & result & Manhattan & learned heuristic (our method) \\ \hline
\multirow{5}{*}{wind flow} & total number of test case & 1284722 & 1284722 \\
& average nodes explored & 552.7 & 265.3 \\
& average time cost (ms) & 63.72 & 22.3 \\ 
& average path length & 17.41 & 17.41 \\
& faster test case ratio & 4.4\% & 95.6\% \\

\hline
\hline
\end{tabular}
\end{table}

\subsection{Result discussion}
The learned heuristic shows superior performance in wind flow risk maps compared to random risk maps. This might be because wind flow risk maps exhibit similar patterns, making them more learnable by neural networks. 
Meanwhile, the results indicate that the learned heuristic reduces the time cost more significantly than the number of nodes explored. With an increase in explored nodes, the list of pending nodes for exploration also grows. Consequently, the time needed to modify this list rises exponentially. This gap could be diminished with further optimization of the data structure for the A* algorithm.


\section{Conclusion}\label{sec: conc}
This study presents a learning-based approach to risk-aware UAV path planning. By integrating transformer-based neural networks with the A* algorithm, we have successfully developed a method that not only adheres to safety constraints but also enhances computational efficiency. Our methodology, evidenced by substantial improvements in node exploration and time savings, represents a significant leap in solving the CSP problem in $16 \times 16$ grid world. The results underscore the effectiveness of our custom transformer-based heuristic, which notably surpasses traditional methods in both speed and efficiency, without compromising the accuracy of the optimal path. This work not only contributes to the field of UAV navigation but also opens new avenues for applying advanced neural network architectures to NP-hard problems.

\section*{Acknowledgments}
\medskip
This work was partially supported by the National Science Foundation under Grants CMMI-2138612. Any opinions, findings, conclusions, or recommendations expressed in this paper are those of the authors and do not reflect the views of NSF.

\bibliographystyle{aiaa}
\bibliography{references}

\end{document}